# Predicting Risk of Dementia with Survival Machine Learning and Statistical Methods: Results on the English Longitudinal Study of Ageing Cohort

Daniel Stamate[1,2], Henry Musto[1,*], Olesya Ajnakina[3,4], Daniel Stahl[3]

1. Data Science & Soft Computing Lab, Computing Department,
   Goldsmiths College, University of London, United Kingdom
   * Joint first-author

2. Division of Population Health, Health Services Research & Primary Care,
   School of Health Sciences, University of Manchester, United Kingdom

3. Institute of Psychiatry Psychology and Neuroscience,
   Biostatistics and Health Informatics Department,
   King's College London, United Kingdom

4. Department of Behavioural Science and Health,
   Institute of Epidemiology and Health Care,
   University College London, United Kingdom

**Abstract.** Machine learning models that aim to predict dementia onset usually follow the classification methodology ignoring the time until an event happens. This study presents an alternative, using survival analysis within the context of machine learning techniques. Two survival method extensions based on machine learning algorithms of Random Forest and Elastic Net are applied to train, optimise, and validate predictive models based on the English Longitudinal Study of Ageing – ELSA cohort. The two survival machine learning models are compared with the conventional statistical Cox proportional hazard model, proving their superior predictive capability and stability on the ELSA data, as demonstrated by computationally intensive procedures such as nested cross-validation and Monte Carlo validation. This study is the first to apply survival machine learning to the ELSA data, and demonstrates in this case the superiority of AI based predictive modelling approaches over the widely employed Cox statistical approach in survival analysis. Implications, methodological considerations, and future research directions are discussed.

## 1    Introduction

Dementia, of which approximately two-thirds constitute Alzheimer's Disease (AD) cases [1], is associated with a progressive decline of brain functioning, leading to a significant loss of autonomy, reduced quality of life and a shortened life expectancy [2]. Accumulated evidence indicates that individuals who have dementia have an excess mortality [3] and a shorter life expectancy [4] than individuals without this disease [5]. In England, dementia is now reported as being the leading cause of death for women, having overtaken cancer and cardiovascular disease [6].

The development of prognostic prediction models, built on combined effects of thoroughly validated predictors, using Machine Learning tools, can be used to forecast the probability of dementia developing within an individual. It is hoped that the availability of such prediction models will facilitate more rapid identification of individuals who are at a higher risk of dementia before the full illness onset [10]. This, in turn, would reduce time to treatment initiation, subsequently minimising the social and functional disability and thereby improving the quality of life for many people affected by these disorders. Identifying individuals at risk of developing dementia would allow the recruiting of patients at high risk for future clinical trials, thereby catalysing the assessment of new treatment or prevention programmes. Furthermore, identifying modifiable risk factors would allow the development of new prevention programmes. For example, there are already some indications that being physically active, staying mentally and socially active, and controlling high blood pressure can potentially deter onset of dementia in the general population [7].

## 2     Literature review

To date, several papers have been published which seek to predict, in a binomial or multinomial classification setting, the probability that any one individual may develop dementia, using neuropsychological test scores, cerebrospinal fluid biomarkers, genetic information, neuroimaging, and demographics data within a fixed period of time. For a recent review, see [8]. These include several studies using the longitudinal Alzheimer's Disease Neuroimaging Initiative (ADNI) study. For example [9] compared several Machine Learning techniques to explore variables found in the ADNI dataset and their suitability as indicator of dementia onset. Although a good performance was demonstrated across all examined algorithms, the best model was Gradient Boosting Machine, with an internally validated area under the curve (AUC) of 0.87. On the other hand, [11] achieved a discriminative accuracy of 0.91 when using ADNI data and support vector machines (SVM) to predict dementia onset. [10] proposed an efficient prediction modelling approach to the risk of dementia based mainly on the Gradient Boosting Machines method, using a large dataset from CPRD (Clinical Practice Research Datalink) repository with data from primary care practices across UK [27], and achieving an AUC performance of 0.83.

Although the ADNI and similar longitudinal studies have as their strength a rich and varied data, the overreliance of the predictive community on these data sources has led to disappointing results when attempting to validate these models on external datasets. The problem is further compounded by the handling of the temporal aspect of the longitudinal studies. Studies that have stuck to the classification-based methodology have dealt with the temporal aspect by including time (defined as discrete time intervals since the start of the study) as a predictor in the model, rather than the outcome of interest [9]. However, by not attempting to predict these temporal aspects, we lose the opportunity to gain clinically relevant information on the expected time to a dementia diagnosis. A common problem of longitudinal studies is drop-out over time where only partial information on a person's survival time is available (Censoring). Furthermore, the standard classification approach is susceptible to instability and inaccuracy when dealing with imbalanced data. Because most subjects do not go on to develop the disease (in this case, dementia), imbalance in the classification outcome must be addressed, usually with under-sampling, over-sampling, or bootstrapping. Such approaches add further complications to a model and the interpretability of its predictions.

A possible solution is one that has in general been less explored so far within the realm of machine learning. This solution is the use of survival methodology as a tool for accounting for and predicting the temporal dynamics of receiving a diagnosis of dementia. In other words, one would seek to utilise the well-established survival techniques found in Cox Proportional Hazards or similar and build upon these frequentist approaches using modified Machine Learning tools [1]. Such an approach would preserve the potential information contained within a temporal outcome, and associated dichotomy of dementia versus no dementia whilst also strengthening the predictive power of the existing frequentist approach by overlaying modified machine learning techniques. Furthermore, it can provide an opportunity to introduce high dimensional data of the type likely to be found when predicting using clinical data. The standard Cox model struggles when confronted with such data, and thus it would be of significant benefit to clinical research if the two approaches could be combined. Finally, survival methods can provide a way to account for censored data whereby subjects are dropping out during the study or are surviving beyond study length. Thus, it can create models which are more robust than standard classification models.

Despite the scarcity of survival modelling papers in relation to dementia prediction, recent examples have shown promise in attempting to outperform the classic Cox proportional hazard model, using survival machine learning and survival deep learning on clinical datasets [12-14]. A pertinent study within the current field of interest is [15] whose authors sought to look at survival machine learning performance when applied to datasets designed for dementia investigation. They found that all machine learning models outperformed the standard Cox Proportional Hazard model. This study, along with those mentioned above, provides support for survival machine learning as a predictive tool for clinical temporal problems.

## 3    Methodology

This paper builds upon [16] which looked at the English Longitudinal Study of Aging (ELSA) and used an accelerated failure time (AFT) survival modelling approach to predicting the time to a subject's likely diagnosis with all-cause dementia. This work found strong evidence that certain features related to socioeconomic markers and genetics play a key role in predicting time to dementia diagnosis.

### 3.1    Problem definition

In this work we propose an approach to predicting the time to a dementia diagnosis, based on survival machine learning techniques such as Survival Random

Forests and Survival Elastic Net, and on a conventional statistical method such as Cox Proportional Hazard model. In order to obtain insight into the predictions, we created and assessed variable importance rankings derived from our best model in this study, which could ideally provide actionable advice for prevention.

### 3.2    Data description

Data was drawn from the English Longitudinal Study of Ageing (ELSA) study, which is a nationally representative sample of the English population aged ≥ 50 years [17]. The ELSA study started in 2002 (wave 1), with participants recruited from an annual cross-sectional survey of households who were then followed up every two years until 2016. Comparisons of ELSA with the national census showed that the baseline sample was representative of the non-institutionalised general population aged 50 and above in the United Kingdom. Ethical approval for each of the ELSA waves (1-8) was granted by the National Research Ethics Service (London Multicenter Research Ethics Committee). All participants gave informed consent. In total, the dataset contained 7556 participants, 45% of which were male.

### 3.3    Ascertainment of dementia cases

To ascertain dementia cases, we used methods with validated utility in population-based cohorts [18-20]. Dementia diagnosis was ascertained at each wave using self-report participant's physician diagnosis of dementia or AD. For those ELSA participants who were unable to respond to the main interview themselves, the Informant Questionnaire on Cognitive Decline in the Elderly (IQCODE) was administered with a score above the threshold of 3.386 indicating the presence of dementia [21-22]; the selected threshold demonstrated both excellent specificity (0.84) and sensitivity (0.82) for detection of all-cause (undifferentiated) dementia [23]. Overall, 83.5% of dementia cases were identified from reports of physician-diagnosed dementia or AD and 16.5% were identified based on the IQCODE score.

### 3.4    Predictors

N = 197 predictor variables related to participants' general health, comorbid health conditions, mental health, cognitive domains, life satisfaction, mobility, physical activity, social-economic status, and social relationships were considered for the model development. The gene APOEe4, a predictor with a well-established link to Alzheimer's risk, was also included as a predictor. For further details see [17].

### 3.5    Data pre-processing

The process of model development, evaluation and validation was carried out according to methodological guidelines outlined by [24]; results were reported according to the Transparent Reporting of a multivariable prediction model for Individual Prognosis Or Diagnosis (TRIPOD) guidelines [25]. Boolean variables were created, indicating the location of missing data for each predictor. All synonyms for missing values were standardised, and duplicate variables were removed. Variables with missingness at 51% or greater of the total rows for that predictor were removed. The cut-off of 51% allowed to include the APOEe4 (a predictor with a well-established link to Alzheimer's risk), which had 50.8% missingness. The remaining predictors had a mean percentage missingness of 1.27%, with a range of 0-50.8%. Missing values were imputed using K-nearest neighbour with K = 5. The data was centered and scaled as part of this process.

We used two versions of the dataset on which we developed our models. The first data version excluded variable scfru which was based on a questionnaire regarding diet and particularly on evaluating a score based on fruit consumption, and the second version included this variable. Variable scrfu was among the variables that showed predictive capability, but also did its NA indicator which in certain cases as dementia may be related to the limited capacity of certain patients to respond to the questionnaire. For this reason, on one hand, we wanted to see the impact of including or excluding scfru in/from our predictive models, and on another hand we compared our models mainly using the performances on the dataset without scfru.

### 3.6    Model development

A simple Cox Proportional Hazard Model (hereafter denoted simply by Cox) was constructed, which served as the baseline for comparison with two survival machine learning models:

1. Cox Penalised Regression using Elastic Net (hereafter denoted simply by ElasticNet) [15], which is similar to the base Cox Proportional Hazard Model but with Elastic Net regularisation, allows the model to shrink the coefficients of less important variables, and even to make them equal to 0, depending on the shrinkage strength and the proportion of the Lasso component in this model. This helps improving prediction accuracy and model interpretability. The main hyperparameters of the model that were tuned were *alpha*, which controls the proportion between the L1 (Lasso) and L2 (Ridge) regularisations, and *lambda*, which controls the strength of the shrinkage. In our tuning grid, the values for *alpha* varied between 0 (corresponding to Ridge regularisation) and 1 (corresponding to Lasso regularisation), with a step of 0.05, while the values for *lambda* varied between 0.05 and 0.3, with a step of 0.05.

2. Survival Random Forest (hereafter denoted simply by RF) [15], is based on the Random Forests algorithm which produces a model formed of an ensemble of trees, each of which learnt on a bootstrap copy of the training set and in the node of which a random sample of predictors of fixed size *mtry*, compete to be selected, with their best split point in order to maximise the survival difference between subsequent nodes [11]. RF has been chosen in this study for its flexibility to capture non-linear patterns in data. Apart from the hyperparameter *mtry* explained above, we used also a hyperparameter called *min.node.size* implementing a pre-pruning criteria for the trees in the RF model to have a minimum number of instances in the terminal nodes. In the tuning grid, the values of *mtry* varied between 10 and the half of the number of columns in the dataset, with a step of 3, while the values for *min.node.size* in the grid were 1, 10, and 20. RF comprised 500 trees which is the default value. The number of trees promotes model convergence (large is better), and in general is not a hyperparameter to tune.

Model tuning was performed using 5-fold cross-validation on the training data set, as part of a nested cross-validation procedure explained below.

### 3.7 Model optimisation and evaluation with nested cross-validation and Monte Carlo validation

A Nested Cross-Validation (NCV) procedure was implemented to tune and evaluate our models with precise estimates of the models' performance. NCV consisted of an outer 3-fold CV, and an inner 5-fold CV.

In order to reliably assess the models' stability, we conducted a Monte Carlo validation procedure (MC), consisting of 90 experiments per model. In each experiment, the dataset was randomly split in 2 thirds for the training data on which the models were tuned with a 5-fold CV, and 1 third for the testing set on which the models were evaluated.

To ensure representativeness of training and test samples in both procedures, NCV and MC, the data splitting was stratified based on the dementia cases variable.

### 3.8 Performance metric

We used the concordance index, called also Harrell's C-index [26] and simply denoted *cindex* here, to assess and compare the prediction performance of the different models. C-index is a generalisation of the ROC AUC metric, and intuitively gives the probability that a predicted risk for dementia is higher for patients with a shorter time to event. More precisely:

$$cindex = C/\ (C+D)$$

where $C$ represents the number of concordant pairs of patients, and $D$ represents the number of discordant pairs of patients [26]; *cindex* is a number between 0 and 1, where 0.5 signifies a random prediction, and 1 indicates that larger times to event concord perfectly with smaller predicted risks.

### 3.9 Software and hardware

The data analysis was conducted using the R language. The stratified data splitting, the KNN imputation and data normalisation via centring and scaling were performed using the Caret R package. The Cox, ElasticNet, and RF survival models were all trained and tuned and evaluated under the umbrella of the MLR3 R package. The hardware consisted of 3 servers running Linux, with Intel 10 cores, AMD Ryzen 16 cores and AMD Ryzen 12 cores, and with 128GB, 128GB and 64GB of RAM, respectively, which were used in our analyses including the computationally intensive tasks for tuning the models, and NCV and MC validation procedures for assessing the models' performances and their stability.

## 4    Results

### 4.1    Internal validation using nested cross-validation

The nested cross-validation cindex performance for train, inner cross-validation, and test or outer cross-validation, for each type of model are detailed below.

**Table 1.** Nested cross-validation results, based on the data without and with scfru variable

| Survival Model | Outer-CV (test) cindex | Train cindex | Inner-CV cindex |
|---|---|---|---|
| Cox +scfru | **0.776** **0.791** | 0.814 0.828 | NA (not tuned) |
| ElasticNet + scfru | **0.843** **0.861** | 0.855 0.873 | 0.840 0.861 |
| RF +scfru | **0.851** **0.867** | 0.972 0.955 | 0.848 0.864 |

The best performing model in terms of the nested cross-validation was Survival Random Forest, followed by Survival Elastic Net, followed by Cox PH model. Hence both machine learning models, RF and ElasticNet, outperformed the conventional statistical model Cox on the test set.

### 4.2    Monte Carlo validation

The results for the Monte Carlo validation are outlined in Table 2 and Figure 1 below.

**Table 2:** Monte Carlo validation results (90 experiments) for Cox, Survival Elastic Net and Survival Random Forest based on the data without and with scfru variable.

| Survival Model | Test cindex mean(SD) | Train cindex mean(SD) | CV cindex mean(SD) |
|---|---|---|---|
| Cox +scfru | **0.761**(0.03) **0.778**(0.034) | 0.793(0.032) 0.807(0.036) | NA (not tuned) |
| ElasticNet + scfru | **0.842**(0.011) **0.862**(0.01) | 0.856(0.004) 0.873(0.004) | 0.841(0.005) 0.861(0.005) |
| RF +scfru | **0.849**(0.009) **0.866**(0.009) | 0.966(0.01) 0.962(0.009) | 0.850(0.005) 0.866(0.005) |

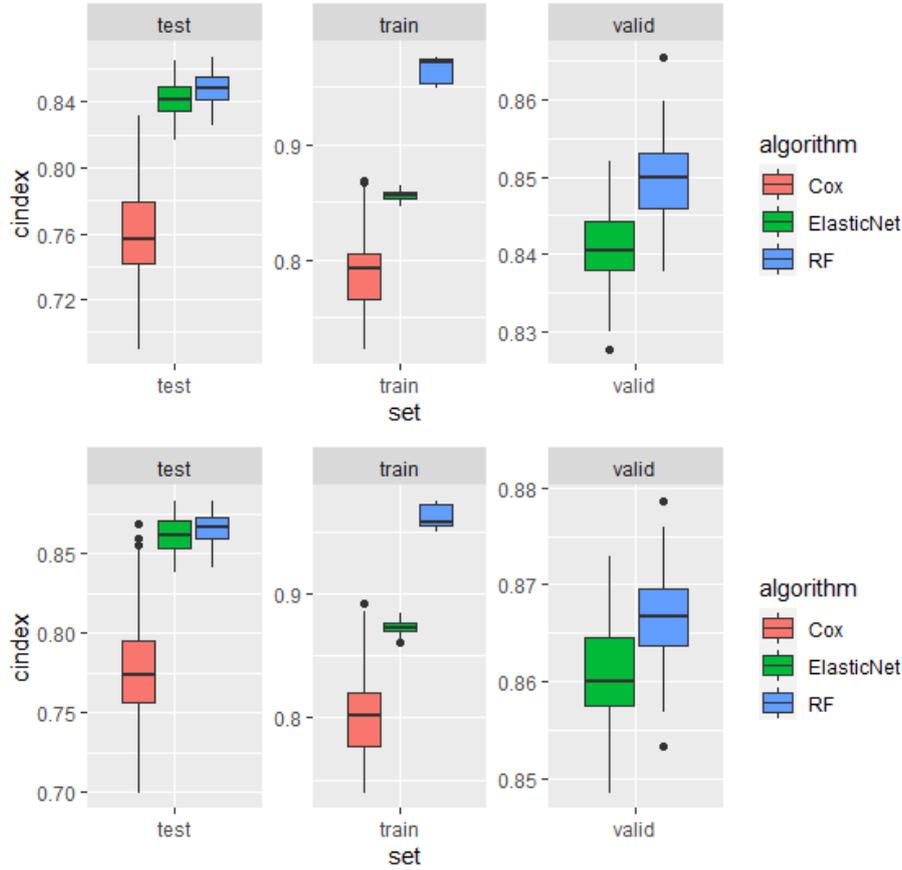

**Fig. 1.** Boxplots for the Monte Carlo derived cindex performances for Cox, Survival Elastic Net and Survival Random Forest. On top: results on dataset without scfru variable, and on bottom: results with scfru variable.

The results in the Monte Carlo validation reveal the following aspects: (a) the machine learning survival models based on Random Forests and Elastic Net demonstrate clearly better mean cindex on the test sets than the conventional statistical model Cox; (b) the results are close to and confirm the estimated performances obtained in the nested cross-validation in Table 1; (c) the standard deviations (provided in brackets in Table 2) for cindex performances on the test sets for Survival Random Forest and Survival Elastic Net are small, and about 3 times smaller than the standard deviations for Cox, which means that the machine learning models are very stable, and by far more stable than the conventional statistical model. This interpretation is confirmed also visually by the boxplots in Fig. 1.

### 4.3 Feature importance

The variable importance computed by Survival Random Forest is provided below. Age was by far the most important variable for the model when predicting time to dementia diagnosis. The other variables in top 20 as importance regard processing speed, self-reported number of hours sleep subjects got in a night, sleep measures (heslpa, heslpd, heslpb, heslpe, heslpf, and headlco), APOEe4, the aggregate measure of memory, and executive functioning. The model also highlighted the role of wealth, social isolation (dhnch, scscc, loneliness_w2, ffamily_w2, and r1retemp) in predicting time to dementia diagnosis.

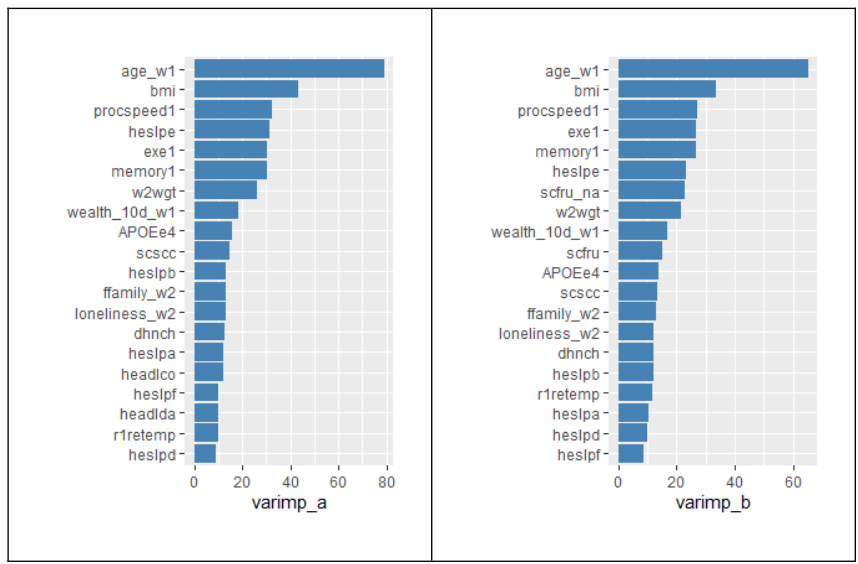

**Fig. 2.** Variable importance for Survival Random Forest (RF) model: on left for dataset without scfru, and on right for data with the scfru variable.

## 5    Discussion

To our knowledge, this paper is the first attempt to develop, evaluate and validate a prediction model for estimating an individual risk of dementia onset in the ELSA dataset using survival machine learning. Our results showed that the machine learning models herein were able to outperform the classic Cox model, with the best performing machine learning technique being the Survival Random Forest followed closely by Survival Elastic Net, as per test results in the nested cross-validation included in Table 1, and test results in the Monte Carlo validation included in Table 2, and Figure 1. Monte Carlo clearly demonstrates the high stability of the survival machine learning models as illustrated by the same Table 2 and Figure 1. The Survival Random Forest model achieved a mean cindex for the test dataset of 0.849 and a standard deviation of 0.009 in 90 Monte Carlo iterations. Survival Elastic Net achieved a mean cindex for the test dataset of 0.842 and a standard deviation of 0.011. Both machine learning models outperformed and were more stable than the Cox model which achieved a mean cindex of 0.761 and a standard deviation of 0.03 (about 3 times larger than the machine learning models) in the Monte Carlo validation. This study indicates strong evidence of machine learning's utility in the field of survival prediction. As mentioned previously, the addition of machine learning paradigms to the classic frequentist survival approaches allows for more variables to be explored than would be possible in a standard Cox proportional hazard model. Moreover, as this study demonstrates, the best survival model based on Random Forests not only improved the predictive accuracy and stability but also provided a useful mechanism to infer the variables' importance, which concords with clinical interpretations of the role of the variables in dementia onset.

In this study, we used two versions of the dataset on which we developed our models. The first data version excluded variable scfru which was based on a questionnaire regarding diet and particularly on evaluating a score based on fruit consumption, and the second version included this variable. The comparisons between the models on the results without or with the scfru variable, lead to the same conclusions in terms of the ranking we established for these methods. Moreover, there is a slight performance increase for all the models on the dataset with the scfru variable, which is the reason why we included these results here. This variable made it in top 20 most important variables, but also did its NA indicator, which in certain cases as dementia, may be related to the limited capacity of certain patients to respond to the questionnaire. For this reason, we compared the three models we developed mainly using the performances on the dataset without scfru.

Although this paper presents examples of good predictive survival machine learning modelling, there are some limitations to this work. Firstly, the data contained predictors with a high percentage of missing values. Although every effort was taken to account for missingness and preserve the pattern of missingness before imputation was performed, a complete dataset may provide results that differ from this work especially if missingness is related to the outcome (not missing at random). Even though dementia and AD were ascertained using a combined algorithm based on a physician made diagnosis and a higher score on the informant reports (IQCODE), it is still reliant on a self-reported diagnosis reported by either the participants themselves or their carers and render more severe cases. Thus, we cannot exclude a possibility that some participants within the "dementia-free" group may have been the preclinical stages of dementia and who, if followed for long enough, might eventually develop dementia. Further, the ELSA dataset is a centre-based data collection study and, although extensive and varied data collection was carried out to try and account for confounding variables, it is possible that other predictors, unmeasured by the data collection procedure, could have an impact on model performance. It is therefore imperative that future work validate these models on different datasets such that the results can be well substantiated. Finally, the ELSA data uses English subjects, who were chosen because they were deemed representative of the United Kingdom at large. Therefore, these results cannot be generalised to populations in other countries. Once again, work must be done to ensure that these results are substantiated by data from subjects in differing datasets.

# 6     Conclusion

This paper represents a first attempt at applying survival machine learning techniques to the ELSA dataset. The intention of this work was to build and validate models which demonstrated good predictive ability on this dataset, specifically in relation to the time to dementia onset. Future work should seek to validate the findings here on other datasets that share similar predictors and outcomes. If the results are substantiated, this could prove to be a new and fruitful approach to clinical prediction modelling of dementia. Another future work will investigate the applicability of an adapted version of the survival machine learning approach we developed here, to the prediction of dementia risk using routine primary care records such as CPRD [27], by extending the machine learning based framework we introduced in [10].

**Acknowledgments**. Daniel Stamate is part-funded by Alzheimer's Research UK (ARUK-PRRF2017-012), the University of Manchester, and Goldsmiths College, University of London. Daniel Stahl is part-funded by the National Institute for Health Research (NIHR) Maudsley Biomedical Research Centre at South London and Maudsley NHS Foundation Trust and King's College London. The views expressed are those of the author(s) and not necessarily those of the NHS, the NIHR or the Department of Health and Social Care. Olesya Ajnakina is further funded by an NIHR Post-Doctoral Fellowship (PDF-2018-11-ST2-020).